\documentclass[10pt,twocolumn,letterpaper]{article}

\usepackage{cvpr}              % To produce the CAMERA-READY version
\usepackage{graphicx}
\usepackage{amsmath}
\usepackage{amssymb}
\usepackage{booktabs}

\newcommand{\cut}[1]{}
\newcommand{\keypoint}[1]{\vspace{0.01cm}\noindent\textbf{#1}}

\usepackage{pifont}
\newcommand{\cmark}{\ding{51}}
\newcommand{\xmark}{\ding{55}}

\usepackage{multirow}
\usepackage[dvipsnames]{xcolor, colortbl}
\usepackage[pagebackref,breaklinks,colorlinks]{hyperref}

\usepackage[capitalize]{cleveref}
\crefname{section}{Sec.}{Secs.}
\Crefname{section}{Section}{Sections}
\Crefname{table}{Table}{Tables}
\crefname{table}{Tab.}{Tabs.}

\def\cvprPaperID{9052} % *** Enter the CVPR Paper ID here
\def\confName{CVPR}
\def\confYear{2023}

\begin{document}

\title{Zero-Shot Everything Sketch-Based Image Retrieval, and in Explainable Style}

\author{Fengyin Lin$^{1*}$ \quad Mingkang Li$^{1*}$ \quad Da Li$^{2\dagger}$ \quad Timothy Hospedales$^{2,3}$ \quad Yi-Zhe Song$^{4}$ \quad Yonggang Qi$^{1}$\\
$^{1}$Beijing University of Posts and Telecommunications \quad $^{2}$Samsung AI Centre, Cambridge\\ $^{3}$University of Edinburgh \quad $^{4}$SketchX, CVSSP, University of Surrey\\
{\tt\footnotesize \{fylin,lmk,qiyg\}@bupt.edu.cn\quad dali.academic@gmail.com\quad t.hospedales@ed.ac.uk\quad y.song@surrey.ac.uk}
% \and
}

\maketitle

%%%%%%%%% ABSTRACT
\begin{abstract}
This paper studies the problem of zero-short sketch-based image retrieval (ZS-SBIR), however with two significant differentiators to prior art (i) we tackle all variants (inter-category, intra-category, and cross datasets) of ZS-SBIR with just one network (``everything''), and (ii) we would \underline{really} like to understand how this sketch-photo matching operates (``explainable''). Our key innovation lies with the realization that such a cross-modal matching problem could be reduced to comparisons of groups of key local patches --  akin to the seasoned ``bag-of-words'' paradigm. Just with this change, we are able to achieve both of the aforementioned goals, with the added benefit of no longer requiring external semantic knowledge. Technically, ours is a transformer-based cross-modal network, with three novel components (i) a self-attention module with a learnable tokenizer to produce visual tokens that correspond to the most informative local regions, (ii) a cross-attention module to compute local correspondences between the visual tokens across two modalities, and finally (iii) a kernel-based relation network to assemble local putative matches and produce an overall similarity metric for a sketch-photo pair.
Experiments show ours indeed delivers superior performances across all ZS-SBIR settings. The all important explainable goal is elegantly achieved by visualizing cross-modal token correspondences, and for the first time, via sketch to photo synthesis by universal replacement of all matched photo patches. 
Code and model are available at \url{https://github.com/buptLinfy/ZSE-SBIR}.

\end{abstract}

\renewcommand{\thefootnote}{}
\footnotetext{$*$ Equal contribution.}
\footnotetext{$\dagger$ Corresponding author.}

%%%%%%%%% BODY TEXT
\vspace{-0.5cm}
\section{Introduction}
\label{sec:intro}

\begin{figure}
    \centering
    \includegraphics[width=\linewidth]{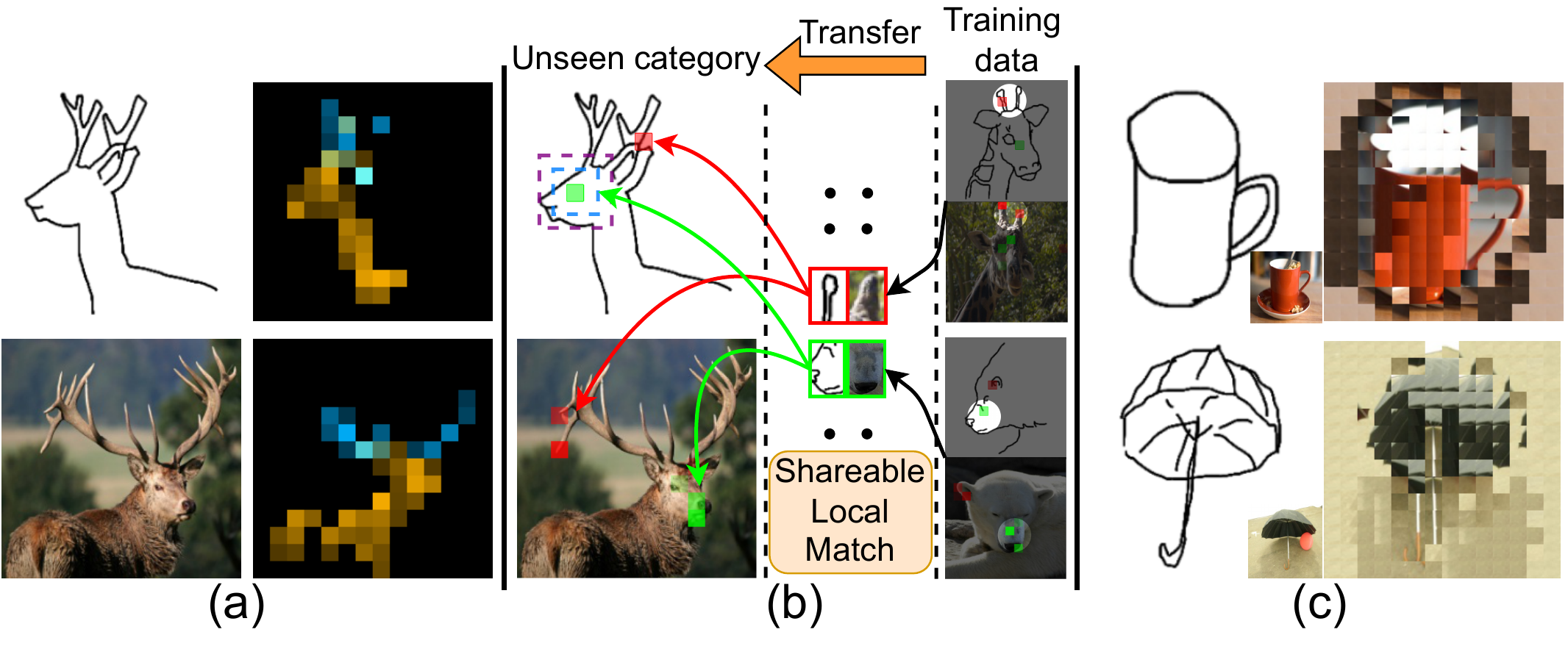}
    \vspace{-0.7cm}
  \caption{Attentive regions of self-/cross-attention and the learned visual correspondence for tackling \emph{unseen} cases. (a) The proposed retrieval token [\texttt{Ret}] can attend to informative regions. Different colors are attention maps from different heads. (b) Cross-attention offers explainability by explicitly constructing local visual correspondence. {The local matches learned from training data are shareable knowledge, which enables ZS-SBIR to work under diverse settings (inter- / intra-category and cross datasets) with just one model. (c) An input sketch can be transformed into its image by the learned correspondence, i.e., sketch patches are replaced by the closest image patches from the retrieved image.}}
  \label{fig:openning}
  \vspace{-0.6cm}
\end{figure}
 
Zero-shot sketch-based image retrieval (ZS-SBIR) is a central problem to sketch understanding~\cite{yang2016zero,kiran2018zero,shen2018zero,dey2019doodle,liu2019semantic,dutta2019semantically,zhang2020zero,dutta2020styleguide,wang2021domain,chaudhuri2022bda,tursun2022efficient,tian2022tvt}. The zero-shot setting is largely driven by the prevailing data scarcity problem of human sketches~\cite{kiran2018zero,dutta2019semantically,xu2022deep} -- they are much harder to acquire compared with photos. As research matures on the non zero-shot~\cite{cao2011edgel,hu2013performance,saavedra2014sketch,parui2014similarity,saavedra2015sketch,yu2016sketch,liu2017deep,xu2018sketchmate}, and in a push to make sketch-based retrieval commercially viable, recent research efforts had mainly focused on ZS-SBIR (or the simpler few-shot setting) \cite{yang2016zero,kiran2018zero,shen2018zero,dey2019doodle,liu2019semantic,dutta2019semantically,wang2021domain}. 

Great strides have been made but attempts have largely aligned with the larger photo-based zero-shot literature, where the key lies in leveraging external knowledge for cross-category adaptation~\cite{dutta2019semantically,liu2019semantic}. That of conducting cross-modal matching is, however, less studied, and most prior art relies on a gold standard triplet loss with some auxiliary modules~\cite{dey2019doodle} to learn a joint embedding. Furthermore, as problems such as domain shift and fine-grained matching come to play, research efforts are mostly done in silo for different settings: category-level (standard)~\cite{yang2016zero,kiran2018zero,shen2018zero}, fine-grained~\cite{bhunia2020sketch}, and cross-dataset~\cite{pang2019generalising}. Last but definitely not least, one can not help but wonder why many of the proposed algorithm work -- what is matched, and how is the transfer conducted?

This paper aims to tackle all said problems associated with the current status quo for ZS-SBIR. In particular, we advocate for (i) a single model to tackle all three settings of ZS-SBIR, (ii) ditching the requirement on external knowledge to conduct category transfer, and more importantly, (iii) a way to explain why our model works (or not).

At the very core of our contribution lies with a well-explored insight that predates ``deep vision'', that image matching can be achieved by establishing local patch correspondences and computing a distance score based on that -- yes, \textit{loosely} similar to that of ``bag of visual words''~\cite{sivic2003video,csurka2004visual,kato2014image} (without building dictionaries). In our context of ZS-SBIR, we would like to conduct this matching (i) cross two diverse modalities in sketch and photo, and (ii) cross-category, granularity (fine-grained or not) and dataset (domain difference) boundaries. The biggest upside of this realization is that just as how ``bag of visual words'' is explainable, we can directly visualize the patch correspondences to achieve a similar level of explainability (see Figure~\ref{fig:openning}).

Our solution first is a transformer-based cross-modal network, that (i) sources local patches independently in each modality, (ii) establishes patch-to-patch correspondences across two modalities, and (iii) computes matching scores based on putative correspondences. We put forward a novel design for each of the three components. We approach (i) by proposing  a novel CNN-based learnable tokenizer, that is specifically tailored to sketch data. This is because the vanilla non-overlapping patch-wise tokenization proposed in ViT~\cite{dosovitskiy2020image} is not friendly to the sparse nature of sketches (as most patches would belong to the uninformative blank). Our tokenizer on the other hand attends to a larger receptive field~\cite{luo2016understanding} hence more keen to sketch data. With this tokenizer, visual cues from nearby regions are aggregated when constructing visual tokens, so that structural information is preserved. In the same spirit of class token developed in ViT for image recognition, we introduce a learnable retrieval token to prioritize tokens for cross-modal matching.

\begin{figure*}
  \centering
   \includegraphics[width=\linewidth]{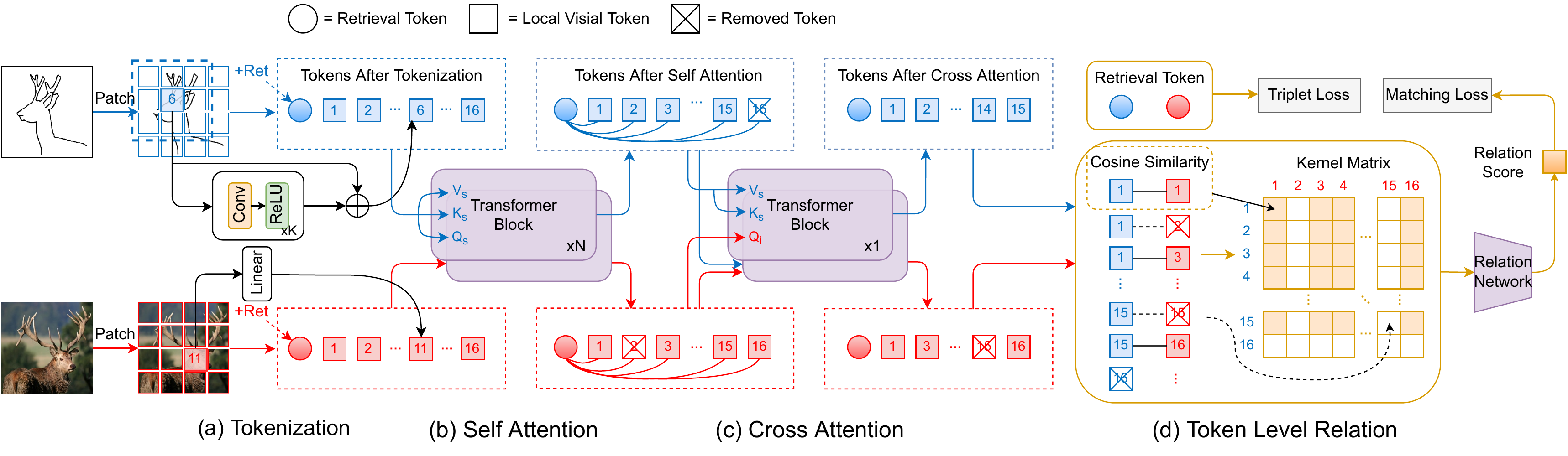}
   \vspace{-0.7cm}
   \caption{Network overview. (a) Learnable tokenization generates structure preserved tokens, preventing the generation of uninformative tokens. (b) Self-attention finds the most informative regions ready for local matching. (c) Cross-attention learns visual correspondence from visual tokens. A retrieval token [\texttt{Ret}] is added as a supervision signal during training. (d) Token-level relation network enables to \emph{explicitly} measure the correspondences of cross-modal token pairs. {Pairs of removed tokens as per token selection will not be counted.}}
   \label{fig:network-overview}
   \vspace{-0.5cm}
\end{figure*}

To establish patch-to-patch correspondences, a novel cross-attention module is proposed that operates across sketch-photo modalities.
Specifically, we propose cross-modal multi-head attention, in which the query embeddings are exchanged between sketch and photo branches to reason patch-level correspondences with only category-level supervision. 
With the putative matches in place, inspired by relation networks~\cite{sung2018learning}, we propose a kernel-based relation network to aggregate the correspondences and calculate a similarity score between each sketch-photo pair.

We achieve state-of-the-art performance across all said ZS-SBIR settings. Explainability is offered (i) as per tradition in terms of visualizing patch correspondences, where interesting local matches can be observed, such as the \texttt{antlers} of \texttt{deer} in Figure~\ref{fig:openning}(b), regardless of a sketch being very abstract, and (ii) by replacing all patches in a sketch with their photo correspondences, to perform sketch to photo synthesis as shown in Figure~\ref{fig:openning}(c).

%------------------------------------------------------------------------
\section{Related Works}

\keypoint{Zero-shot SBIR.} Most previous works \cite{yang2016zero,kiran2018zero,shen2018zero,dey2019doodle,liu2019semantic,dutta2019semantically,zhang2020zero,wang2021domain} treat zero-shot SBIR (ZS-SBIR) as a category-level retrieval problem. Zero-shot learning (ZSL) algorithms \cite{zhang2015zero,changpinyo2016synthesized,bucher2016improving,kodirov2017semantic} typically play a central role in tackling it. The fundamental idea is to map sketch and photo into a shared semantic feature space to help alleviate the cross-domain gap. To assist knowledge transfer to match sketch and photo in unseen categories, side knowledge is normally required, such as text-based class descriptions. 
Essentially, these methods learn to associate sketch and photo to some class-specific feature representations for pairing. However, this matching is limited to a coarse-grained level. The visual similarities in fine details, i.e., visual local correspondences between sketch and photo, are largely ignored during training and inference. More recently, CC-DG \cite{pang2019generalising} formulates ZS-SBIR as a cross-category generalization problem. In particular, CC-DG provides fine-grained SBIR by comparing sketch and photo conditioned on a visual trait that is dynamically selected from a trait bank learned from seen sketches. Although sharing the same goal of fine-grained retrieval, our approach differs from CC-DG that we explicitly learn local visual correspondence between sketch and photo, thereby offering distinct explainability.

\keypoint{Transformer-based cross attention.} Although originating from natural language processing, the transformer \cite{vaswani2017attention,devlin2018bert} has emerged as an effective base network for solving many vision tasks due to its powerful feature representation. Apart from applying self attention, reasoning based on cross attention has been shown effective on image classification, few-shot learning and sketch segment matching. CrossTransformer \cite{doersch2020crosstransformers} finds pixel-level correspondence between a query image and a set of support images for few-shot image classification. The animation transformer (AnT) \cite{casey2021animation} learns segment-level correspondence between human-drawn animations for AI-assisted colorization. Built on vision transformer (ViT) \cite{dosovitskiy2020image}, CrossViT \cite{chen2021crossvit} offers a dual-branch ViT which  extracts multi-scale (small and large image patches) tokens and shares knowledge between two branches by exchanging the class (\texttt{CLS}) tokens, resulting an enhanced \texttt{CLS} token for classification. {Despite sharing the same spirit, our model differs from CrossViT significantly in two ways: CrossViT performs within-branch attention over visual tokens. In contrast, ours carries out cross-branch interaction on visual tokens. Additionally, the output of CrossViT is an augmented \texttt{CLS} token, while ours leverages the visual tokens after cross-attention for local matching by the following token-level relation network.}

\keypoint{Visual correspondence learning.} Learning dense visual correspondence is an essential component and it has been actively explored in many vision tasks, such as structure-from-motion (SfM) \cite{schonberger2016structure}, visual localization \cite{mur2015orb}, simultaneous localization and mapping (SLAM) \cite{davison2007monoslam} and exemplar-based image-to-image translation \cite{zhang2020cross}. Early approaches \cite{lowe2004distinctive,fischler1981random,yi2016lift,ono2018lf} focus on matching hand-crafted features for predicting correspondence between images. Coupled with deep features, plenty of improved results have been achieved. However, most works handle within-domain correspondence between natural images \cite{long2014convnets,choy2016universal,ham2017proposal,han2017scnet,kim2017fcss,aberman2018neural,lee2019sfnet}, or focus on dense pixel-wise correspondence learning across domains from strictly aligned cross-modal data (e.g., the edge map or semantic layout is paired with a sourced photo), and the training objectives are often highly complex (6 or more loss terms) that often requires dedicated engineering \cite{zhang2020cross,zhan2022marginal}. We instead tackle sketch to photo semantic correspondence learning at patch-level, to generalize image matching through local visual evidences.

%------------------------------------------------------------------------
\section{Methodology}

The overall scheme of our proposed framework is shown in~Figure~\ref{fig:network-overview}. Each key module is detailed in the following.

\subsection{Learnable Tokenization}
\label{sec:tokenizer}

Given a query sketch $S \in \mathbb{R}^{h\times w\times c} $ and a gallery image $I \in \mathbb{R}^{h\times w\times c}$ to be matched, we can tokenize them into a sequence of visual tokens by using the same approach proposed in ViT \cite{dosovitskiy2020image} where images are evenly partitioned into non-overlapping patches, followed by a projection head mapping them into $S' \in \mathbb{R}^{n \times d}$ and  $I' \in \mathbb{R}^{n \times d}$.  However, we found that this tokenization is not friendly to sketches, which are typically composed of sparse strokes. To address this issue, we propose a learnable tokenizer, which transforms a given sketch $S \in \mathbb{R}^{h\times w\times c}$ into a sequence of visual embeddings $X \in \mathbb{R}^{n\times d}$. Specifically, the tokenizer is made up of a stack ($K=4$) of convolution layers (Conv) with various kernel size, each followed by a non-linear activation ($\sigma$= ReLU): $X = [\sigma (\mathrm{\texttt{Conv}(S)})]_{\times 4}$.
Essentially, it can enlarge the receptive field when constructing visual tokens through hierarchical convolution, thereby better preserving structural cues from nearby regions.
Additionally, a residual connection is introduced to rectify the vanilla tokens, then the final token embedding is: $X = X + S'$.

\subsection{Self-attention with a Retrieval Token}
Different from the vanilla ViT~\cite{dosovitskiy2020image}, we replace the vanilla class token with a retrieval token $[\texttt{Ret}]$ to facilitate our retrieval task by capturing a global representation of an image. Specifically, the retrieval token is initialized as a trainable $d$-dimensional token embedding $[\texttt{Ret}] \in \mathbb{R}^d$. During the model inference, all visual tokens including the retrieval token $[\texttt{Ret}]$ interact with each other through the multi-head self attention (MSA) modules, followed by MLP blocks, as per~\cite{dosovitskiy2020image}. In formal, the overall forward is as
\begin{align}
    X_0 &= [\texttt{Ret}, X^{1},\dots, X^{n}], \\
    \label{eq:msa}
    X_l &= \text{MSA}(\text{LN}(X_{l-1})) + X_{l-1}, \quad l = 1 \dots L, \\
    \label{eq:mlp}
    X_l &= \text{MLP}(\text{LN}(X_{l})) + X_{l}, \quad l = 1 \dots L, 
\end{align}
where residual connection is introduced in both Eq.~\ref{eq:msa} and~\ref{eq:mlp}. $\text{LN}$ is layer norm and $L$ is the number of layers. The same inference architecture is applied onto photo inputs. Specifically, the MSA module has three different projection heads $[W_{q}, W_{k}, W_{v}]$, which map the same token embeddings into Queries, Keys and Values. Formally, it is formulated as
\begin{equation}
    Q = X^i \cdot W_{q},~~~K = X^i \cdot W_{k},~~~ V= X^i \cdot W_{v}.
\end{equation}
Then the scaled dot-product attention is given by
\begin{equation}
    \texttt{s-attn}(Q,K,V) = \texttt{softmax}(\frac{QK^T}{\sqrt{d}})V.
\end{equation}

\subsection{Cross-modal Attention}
The self-attention module learns an informative token-based representation of each image. To estimate local visual correspondences between sketch and photo tokens, we resort to cross attention. 
The idea is to find pair-wise connections between visual tokens from different modalities, i.e., sketch and photo. This can be achieved by swapping the sketch query $Q_S$ and image query $Q_I$, resulting in the new Query, Key and Value tuples, i.e., $(Q_S, K_I, V_I)$ and $(Q_I, K_S, V_S)$. The cross modal attention is obtained by
\begin{equation}
    \texttt{c-attn}(Q_{I},K_{S},V_{S}) = \texttt{softmax}(\frac{Q_{I}K_{S}^T}{\sqrt{d}})V_{S}.
\end{equation}
In this way, sketch token embeddings are updated by the information from photo tokens. 
Photo token embeddings can be obtained in the same way.

\subsection{Token Selection}
Local visual tokens may represent background or meaningless regions which are unimportant to the retrieval. Thus, a token selection is applied to narrow the scope of attentive tokens, while this reduces computational complexity. Inspired by \cite{liang2022not}, attention scores between the retrieval token \texttt{[Ret]} and all visual tokens are leveraged as a token importance indicator. Formally, attention scores are computed as follows:
\begin{equation}
    a = \texttt{softmax}(\frac{Q_{\texttt{[Ret]}}K^T}{\sqrt{d}}),
\end{equation}
where $a\in \mathbb{R}^n$ and the i-th entry $a^i$ denotes how much information the i-th token contributes to the retrieval token. Consequently, only the top-k visual tokens will be preserved according to the attention scores $a$, and the rest ones are discarded. In practice, token selection is performed at the 4-th, 7-th and 10-th layer in the self-attention, with keep rates $r_S^{SA}$ and $r_I^{SA}$ set for sketch and image respectively at the selected layers. Apart from self-attention, token selection can also be carried out during cross-attention, i.e., using the sketch retrieval token to select image visual tokens with keep rate $r^{CA}$, to prioritize tokens useful for retrieval.

\subsection{A Kernel based Relation Network}
\label{sec:relation}

\keypoint{Cosine kernel matrices generation.} We further introduce a cosine kernel function after the cross attention module to explicitly measure the similarity between each pair of visual tokens. Specifically, given any pair of tokens across two modalities, i.e., $X_S^i$ and $X_I^j$, the kernel matrices $M \in \mathbb{R}^{n\times n}$ is defined as
\begin{equation}
\label{eq:token}
    M^{S,I}_{i, j} = \frac{X_{S}^{i} \cdot {X_{I}^{j}}^T}{\|X_{S}^{i}\| \|X_{I}^{j}\|}.
\end{equation}
This matrix $M$ summarizes the cosine similarity between all pairs of sketch and photo tokens. Importantly, the formed kernel matrix $M$ enables the explicit reasoning on token correspondences by a relation net which is described next.

\keypoint{Relation network.} Inspired by Relation Network proposed in~\cite{sung2018learning}, we incorporate a relation network in our framework to estimate the matching score of a particular sketch-photo pair (S,I), based on their associated local correspondence kernel matrix  $M^{S,I}$. 
Specifically, our relation network $R_{\psi}(\cdot)$ is a stack of two FC-ReLU-Dropout layers that can produce a relation score in the range of $(0,1)$: 
\begin{equation}
\label{eq:retrieval}
\begin{aligned}
    r(S,I) = \texttt{sigmoid}(R_{\psi}(M^{S,I})).
\end{aligned}
\end{equation}
{Unlike concatenating global image features in \cite{sung2018learning}, our relation network conducts reasoning on local token similarities, thereby has the opportunity to learn which (set of) token correspondences (embedded in $M^{S,I}$) to prioritize during matching.} In the end, retrieval can be performed by ranking gallery images according to their relation scores.

\subsection{Losses}
We exploit two losses to train our framework: A triplet loss applied on the $[\texttt{Ret}]$ token, and a regression loss applied on our similarity score $r$. 
Given a triplet $<S_i, I_i^+, I_i^->$, where $S_i$ is an anchor sketch, $I_i^+$ is a photo with the same label to $S_i$ while $I_i^-$ from a different class, the triplet loss is minimized to align the positive pair $<S_i, I_i^+>$, and push the anchor $S_i$ away from the negative instance $I_i^-$. In our case, the retrieval token $[\texttt{Ret}]$ is used as the global feature of sketches and photos, thus the triplet loss is defined as
\begin{equation}
    \begin{aligned}
        \mathcal{L}_{tri} = &\frac{1}{T}\sum_{i=1}^T\max\{||\texttt{Ret}({S_i)} - \texttt{Ret}(I_i^+)|| \\
                            &- ||\texttt{Ret}(S_i) - \texttt{Ret}(I_i^-)||+m, \quad 0\}.
    \end{aligned}
\end{equation}
$T$ is the total number of triplets, and $m$ is the margin. 

In addition, a relation loss is used to measure whether a sketch-photo pair belongs to the same class/instance or not through our kernel based relation network as described in Section~\ref{sec:relation}. Specifically, we regress the predicted relation score $r$ to the ground-truth, i.e., $r = 1$ when matched, and $r = 0$ otherwise. Formally, the matching loss is defined as mean square error (MSE):

\begin{equation}
\label{eq:rn-loss}
    \mathcal{L}_{re} = \sum_{i=1}^N\sum_{j=1}^H (r_{i,j} - \textbf{1}(y_i == y_j))^2,
\end{equation}
$N$ and $H$ are the total numbers of query sketches and candidate photos, respectively. $y$ is the class label. To this end, the overall loss is summed as
\begin{equation}
    \mathcal{L} = \mathcal{L}_{tri} + \mathcal{L}_{re}.
\end{equation}

\subsection{Implementation Details}
Sketch or image is scaled to $224\times 224$. As stated in Section~\ref{sec:tokenizer}, there are four convolution layers in the tokenizer. The kernel size of the first conv-layer is  $7\times 7$ and  $3\times 3$ for the rest. Stride is $2$ for all conv-layers. Consequently, there are 196 visual tokens produced, each is represented by a $d=768$ dimensional vector. The self-attention module is designed as per ViT \cite{dosovitskiy2020image} with 12 blocks and 
pre-trained on ImageNet-1K~\cite{deng2009imagenet}. The cross-attention module is much lighter than the self-attention module and only contains one layer with 8 heads. AdamW is used with lr $10^{-5}$. 

%------------------------------------------------------------------------

\section{Experiments}

To verify the efficacy of our model, we first conduct experiments on category-level ZS-SBIR, followed by an ablation study of key components and explainability analysis to reveal why and how our approach works. Then, we conduct fine-grained SBIR experiments to verify the instance-level retrieval of our proposed model. Finally, cross-dataset ZS-SBIR experiments are conducted to verify the generalization of the learned visual correspondence across different datasets in a zero-shot setting.

\subsection{Category-level ZS-SBIR}\label{cl-zssbir}
\keypoint{Datasets.} There are three large-scale datasets that are commonly used for category-level ZS-SBIR: TU-Berlin Ext \cite{zhang2016sketchnet}, Sketchy Ext \cite{liu2017deep} and QuickDraw Ext \cite{dey2019doodle}. For TU-Berlin Ext, sketches are collected from TU-Berlin dataset \cite{eitz2012humans} which contains 250 categories with 80 sketches per class, and 204,489 photos are sourced from ImageNet \cite{deng2009imagenet} and web images to pair with the sketches. Sketchy Ext is an extended version of Sketchy dataset \cite{sangkloy2016sketchy}, which consists of 125 categories (100 photos per class and 5-8 corresponding sketches per photo). In particular, an additional 60,502 photos are included, resulting in an enlarged photo gallery with 73,002 photos. QuickDraw Ext is the largest SBIR dataset which is composed of 330,000 sketches and 204,000 photos over 110 categories. Following \cite{liu2019semantic}, we split TU-Berlin Ext into 220/30 categories for training/test, and a partition of 100/25 training/test categories  for Sketchy Ext. In addition, a split of 104/21 training/test classes for Sketchy Ext proposed in \cite{kiran2018zero} is also used for evaluation when the testing classes are not presented in the ImageNet-1K classes. The default data split in \cite{dey2019doodle} is applied to QuickDraw Ext, i.e., 80 classes for training and 30 for testing. 

\begin{table*}[t]
\small
\begin{center}
\caption{Category-level ZS-SBIR comparison results.  ``ESI'' : External Semantic Information. ``-'' :  not reported. {The best and second best scores are color-coded in \textcolor{red}{red} and \textcolor{blue}{blue}.}}
\vspace{-0.3cm}
\label{compare}
\label{table:results}
\scalebox{0.84}{
\begin{tabular}{lclccccccccc}
\toprule
\multirow{2}{*}{\bf Method} & \multirow{2}{*}{ESI} & \multirow{2}{*}{{$\mathbb{R}^D$}}  & \multicolumn{2}{c}{\bf TU-Berlin Ext} & \multicolumn{2}{c}{\bf Sketchy Ext} & \multicolumn{2}{c}{\bf Sketchy Ext \cite{kiran2018zero} Split} & \multicolumn{2}{c}{\bf QuickDraw Ext} \\ 
\cline{4-11}
& & & mAP & Prec@100  &  mAP & Prec@100 &  mAP@200 & Prec@200 & mAP & Prec@200 \\
\midrule
{ZSIH \cite{shen2018zero}} & \cmark & 64 & 0.220 & 0.291  & 0.254 & 0.340 & - & - & - & - \\
{CC-DG \cite{pang2019generalising}} & \xmark & 256 & 0.247 & 0.392 & 0.311 & 0.468 & - & - &  - &  - \\
{DOODLE \cite{dey2019doodle}} & \cmark & 256 & 0.109 & -  & 0.369 & - & - & - & 0.075 & 0.068 \\
{SEM-PCYC \cite{dutta2019semantically}} & \cmark & 64 & 0.297 & 0.426 & 0.349 & 0.463 & - & - & - & - \\
{SAKE \cite{liu2019semantic}} & \cmark & 512 & 0.475 & 0.599 & 0.547 & 0.692 & 0.497 & 0.598 & 0.130  & 0.179   \\
{SketchGCN \cite{zhang2020zero}} & \cmark & 300 & 0.324 & 0.505 & 0.382 & 0.538 & - & - & - & - \\
{StyleGuide \cite{dutta2020styleguide}} & \xmark & 200 & 0.254 & 0.355 & 0.376 & 0.484 & 0.358 & 0.400 & - & - \\
{PDFD \cite{deng2020progressive}} & \cmark & 512 & 0.483 & 0.600 & 0.661 & 0.781 & - & - & - & - \\
ViT-Vis \cite{dosovitskiy2020image} & \xmark & 512 & 0.360 & 0.503 & 0.410 & 0.569 & 0.403  & 0.512 & 0.101 & 0.113 \\
ViT-Ret \cite{dosovitskiy2020image} & \xmark & 512 & 0.438 & 0.578 & 0.483 & 0.637 & 0.416  & 0.522 & 0.115 & 0.127 \\
{DSN \cite{wang2021domain}}  & \cmark & 512 & 0.484 & 0.591  & 0.583 & 0.704 & - & - & - & - \\
{BDA-SketRet \cite{chaudhuri2022bda}} & \cmark & 128 & 0.375 & 0.504 & 0.437 & 0.514 & \textcolor{red}{0.556} & 0.458 & \textcolor{red}{0.154} & \textcolor{red}{0.355} \\
{SBTKNet \cite{tursun2022efficient}} & \cmark & 512 & 0.480 & 0.608 & 0.553 & 0.698 & 0.502 & 0.596 & -  & -   \\
{Sketch3T \cite{sain2022sketch3t}} & \cmark & 512 & 0.507 & - & 0.575 & - & - & - & - & - \\
{TVT \cite{tian2022tvt}} & \cmark & 384 &  0.484 & \textcolor{red}{0.662} &  0.648 & 0.796 & \textcolor{blue}{0.531} & \textcolor{blue}{0.618} & \textcolor{blue}{0.149} & \textcolor{blue}{0.293}  \\
\midrule
\rowcolor{lightgray}
Ours-RN  & \xmark & 512 & \textcolor{blue}{0.542} & \textcolor{blue}{0.657} & \textcolor{blue}{0.698} & \textcolor{blue}{0.797} & 0.525 & \textcolor{red}{0.624} & 0.145 & 0.216 \\
\rowcolor{lightgray}
Ours-Ret  & \xmark & 512 & \textcolor{red}{0.569} & {0.637} & \textcolor{red}{0.736} & \textcolor{red}{0.808} & 0.504 & 0.602 & 0.142 & 0.202 \\
\bottomrule
\end{tabular}
}
\end{center}
\end{table*}

\begin{figure*}[h]
\vspace{-0.7cm}
  \centering
%   \fbox{\rule{0pt}{2in} \rule{\linewidth}{0pt}}
   \includegraphics[width=0.8\linewidth]{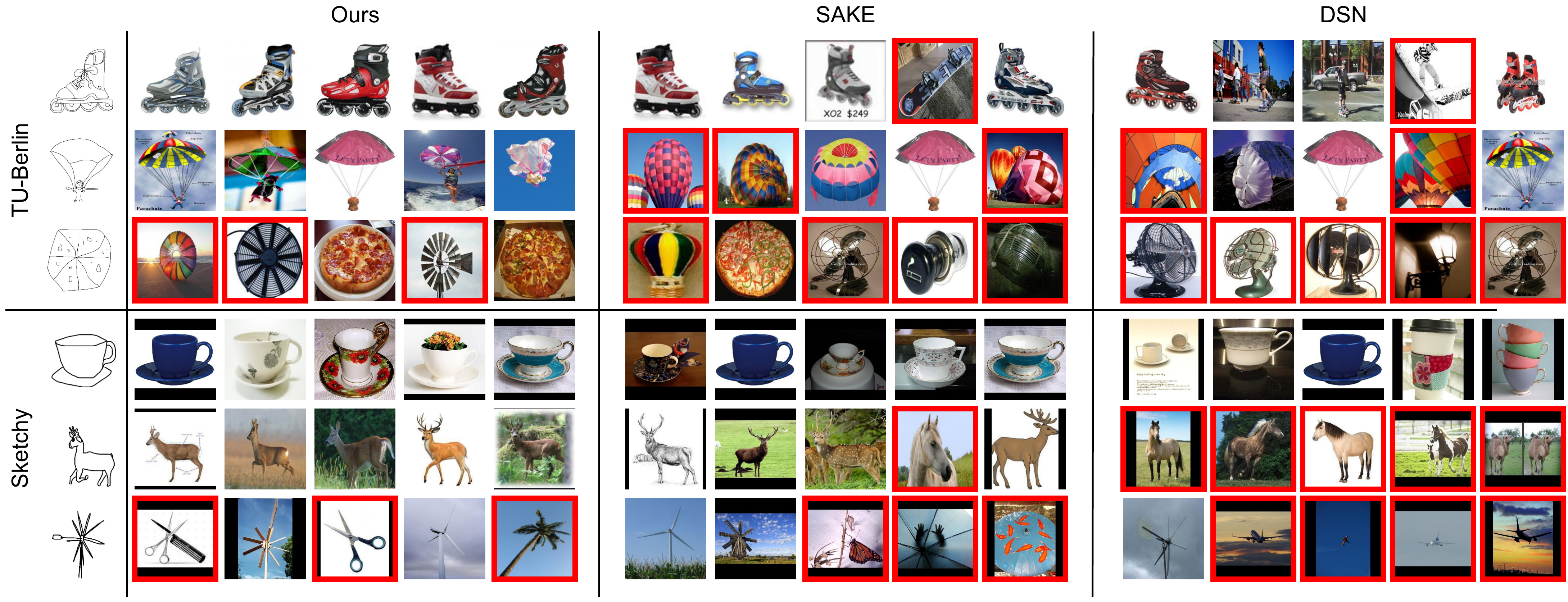}
   \vspace{-0.3cm}
   \caption{Exemplar comparison retrieval results for the given query sketches and the top 5 retrieved images. \textcolor{red}{Red} box denotes false positive.}
   \label{fig:retrieval}
   \vspace{-0.4cm}
\end{figure*}

\begin{table}[t]
\small
\begin{center}
\vspace{-0.2cm}
\caption{Generalized ZS-SBIR results.}
\vspace{-0.4cm}
\label{table:gzs-sbir}
\scalebox{0.78}{
\begin{tabular}{lcccc}
\toprule
\multirow{2}{*}{\bf Method} & \multicolumn{2}{c}{\bf TU-Berlin Ext} & \multicolumn{2}{c}{\bf Sketchy Ext}  \\ 
\cline{2-5}
& mAP & Prec@100  &  mAP & Prec@100  \\
\hline
SEM-PCYC \cite{dutta2019semantically} & 0.192  & 0.298  & 0.307 & 0.364  \\
StyleGuide \cite{dutta2020styleguide} & 0.149  & 0.226  & 0.331 & 0.381  \\
BDA-SketRet \cite{chaudhuri2022bda} & 0.251  & 0.357  & 0.338 & 0.413  \\
SBTKNet \cite{tursun2022efficient} & 0.334  & 0.494  & 0.515 & 0.572  \\
\hline
Ours-RN & 0.432  & 0.460  & 0.634 & 0.651  \\
Ours-Ret & 0.464  & 0.485  & 0.656 & 0.670  \\
\bottomrule
\end{tabular}
}
\end{center}
\vspace{-0.8cm}
\end{table}

\keypoint{Competitors.} We compare our method with several baselines, including ZSIH \cite{shen2018zero}, CC-DG \cite{pang2019generalising}, DOODLE \cite{dey2019doodle}, SEM-PCYC \cite{dutta2019semantically}, SAKE \cite{liu2019semantic}, SketchGCN \cite{zhang2020zero}, StyleGuide \cite{dutta2020styleguide}, PDFD \cite{deng2020progressive}, DSN \cite{wang2021domain}, BDA-SketRet \cite{chaudhuri2022bda}, SBTKNet \cite{tursun2022efficient}, Sketch3T \cite{sain2022sketch3t}, TVT \cite{tian2022tvt} and ViT-Ret/ViT-Vis \cite{dosovitskiy2020image} adapted by us. ViT-Ret means replacing the class token in ViT with a retrieval token used for matching; while ViT-Vis uses the visual tokens for matching.
\cut{ZSIH \cite{shen2018zero} deploys semantic graph convolutional networks to inject semantic relations of training categories into the learning procedure. CC-DG \cite{pang2019generalising} is a domain generalization approach for ZS-SBIR. In particular, it introduces fine-grained visual trait features into the retrieval, a similar motivation to ours. DOODLE \cite{dey2019doodle} enforces the model to encode semantic information by reconstructing Word2Vec \cite{mikolov2013efficient} embeddings, while SEM-PCYC \cite{dutta2019semantically} employs two text-based models (Word2Vec \cite{mikolov2013efficient} and GloVe \cite{pennington2014glove}) as side semantic information for training. And SAKE \cite{liu2019semantic} brings a pre-trained 1,000 class ImageNet classifier as a teacher network, as well as applying WordNet as semantic constraint.}It should be noted that all the baselines, except CC-DG \cite{pang2019generalising}, StyleGuide \cite{dutta2020styleguide} and ViT variants \cite{dosovitskiy2020image}, employ external semantic information, whereas our method only relies on the learned visual correspondences between sketch-photo pairs. 
We also compare two variants of our model, i.e., Ours-Ret and Ours-RN, which retrieve images based on the retrieval token and the relation network separately.

\keypoint{Evaluation protocol.} Mean average precision (mAP), precision on top 100 (Prec@100) and top 200 (Prec@200) are reported following the standard evaluation protocol.

\keypoint{Results.} From Table~\ref{table:results}, we can see that our proposed method achieves better or comparable results over other competitors, which is especially noteworthy since our method does not benefit from extra semantic information, i.e., text or class label.
Meanwhile, we can see Ours-Ret achieves better results on TU-Berlin Ext and Sketchy Ext datasets, and works slightly worse than Ours-RN on Sketchy Ext \cite{kiran2018zero} Split and QuickDraw Ext. Some qualitative results (including some failure cases) are illustrated in Figure~\ref{fig:retrieval}. We can observe that, compared with SAKE and DSN, most of the top  images retrieved using our approach  more faithfully resemble to the query sketches in terms of the overall object pose and shape characteristics. And the false positives given by our method are somehow reasonable, as they are superficially similar to the query. 
{We additionally compare ours with baselines for generalized ZS-SBIR. Results in Table~\ref{table:gzs-sbir} show that ours clearly outperforms others, suggesting a strong generalizability of our model.}

\begin{figure}[t]
  \centering
%   \fbox{\rule{0pt}{2in} \rule{\linewidth}{0pt}}
    \vspace{-0.2cm}
   \includegraphics[width=\linewidth]{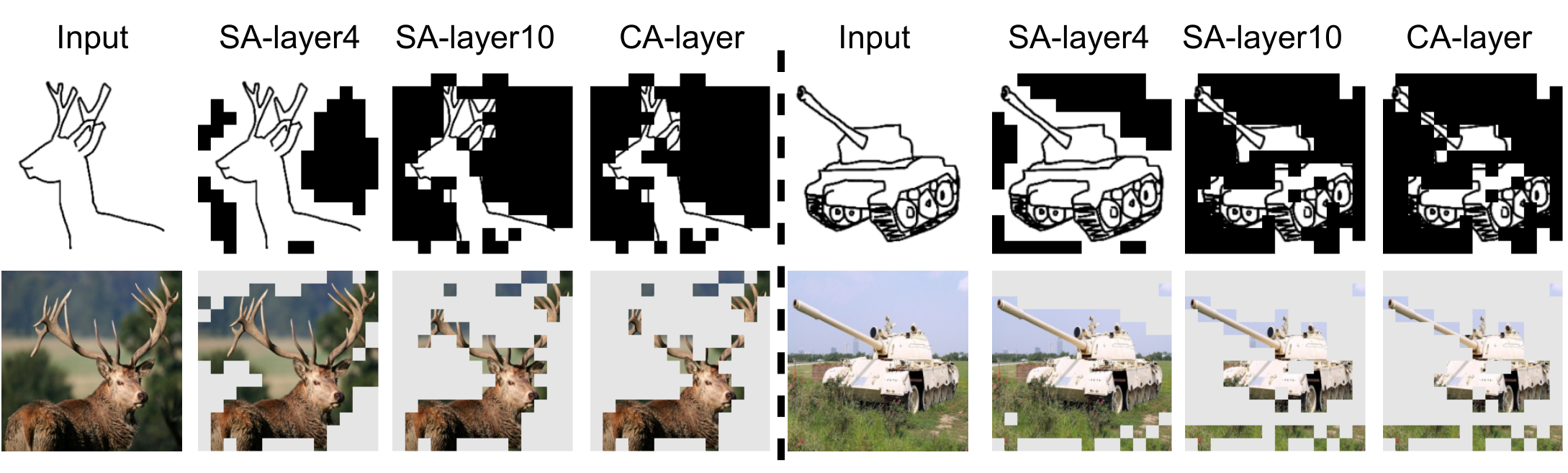}
   \vspace{-0.7cm}
   \caption{Visualization of token selection at different layers by setting keep rate for SA layers to 0.7 and the CA layer to 0.9.}
   \label{fig:token}
   \vspace{-0.4cm}
\end{figure}

\keypoint{Ablation study.} An ablative study is conducted to examine the importance of each key component in our model. In particular, based on Ours-RN (Ours-full), we remove every individual \textbf{\emph{component}} at a time with other parts remaining. Specifically, \textbf{w/o CA}: The cross-attention layers are removed, and the resulting visual tokens from the self-attention module are fed directly into the relation network. \textbf{w/o SA}: Self-attention layers are replaced by ResNet-50 to generate tokens. The sketch/photo feature map given by ResNet-50 is  transformed into a sequence of feature vectors, serving as input tokens to the cross-attention module. \textbf{w/o cosine kernel (Cos-K)}: Instead of calculating the cosine kernel between the visual tokens, we concatenate each pair of them with a learned distance metric, which is inspired by \cite{sangkloy2016sketchy}, for matching. \textbf{w/o RN loss}: {Relation loss in Eq.~(\ref{eq:rn-loss}) is disabled during model training.} \textbf{w/o [\texttt{Ret}]}: [\texttt{Ret}] and triplet loss are discarded, while only the visual tokens are used throughout the model training and testing. \textbf{w/o learnable tokenizer (L-Tok)}: The learned tokenzier is reverted back to the vanilla one, i.e., images are evenly divided into 16x16 patches. 
We can see from Table~\ref{table:abla}: (i) Without self-attention (SA) or cross-attention (CA), the performance drops dramatically, indicating the importance of each component. (ii) Using visual tokens only leads to some performance degradation on TU-Berlin Ext and Sketchy Ext, confirming the effect of using [\texttt{Ret}] token. (iii) Our proposed learnable tokenizer is clearly helpful on both datasets. (iv) Removing the cosine kernel is also harmful, confirming its importance. \textbf{\emph{Token selection}}: Moreover, we employ a recent work~\cite{liang2022evit} to study how the keep rate of tokens in our self-/cross-attention influences the final results. We can see that a lower keep rate will lead to a slight performance drop yet a significant speedup, i.e., speedier runtime per pair matching (RPM). Setting the token keep rate for both sketch and image to 0.7 is the best considering the speed gain versus the performance loss (Table~\ref{table:abla} row in brown). Figure~\ref{fig:token} shows visualization of token selection. 

\setlength{\tabcolsep}{2pt}
\begin{table}[t]
% \small
\centering
\caption{Ablation study results on manifesting importance of each key \emph{component}, and using different \emph{token selection rates}. \cut{``SA'' denotes self-attention module. ``CA'' denotes the cross-attention module.} }
\vspace{-0.3cm}
\label{table:abla}
\resizebox{.85\linewidth}{!}{
\begin{tabular}{llccccccc}
\toprule
& \multirow{2}{*}{\bf Model} & \multicolumn{2}{c}{\bf Keep Rate} & \multicolumn{2}{c}{\bf TU-Berlin Ext} & \multicolumn{2}{c}{\bf Sketchy Ext} & \multirow{2}{*}{\bf RPM} \\ 
\cmidrule{5-8}
& & $r_S^{SA}$/$r_I^{SA}$ & $r^{CA}$ & mAP & Prec@100  &  mAP & Prec@100 & (ms)\\
\midrule
\multirow{6}{*}{\rotatebox{90}{Components}} & w/o CA & - & - & 0.294 & 0.352  & 0.295 & 0.346 & -\\
& w/o SA & - & - & 0.256 & 0.388  & 0.286 & 0.381 & -\\
% \hline
& w/o Cos-K & - & - & 0.342 & 0.419  & 0.390 & 0.481 & -\\
& w/o RN loss & - & - & 0.497 & 0.610  & 0.656 & 0.744 & -\\
& w/o [\texttt{Ret}] & - & - & 0.519 & 0.623  & 0.681 & 0.767 & -\\
% \hline
& w/o L-Tok & - & - & 0.514 & 0.621  & 0.672 & 0.767 & - \\
\midrule
% \rowcolor{lightgray}
 & {Ours-full} & {-/-} & {-} & \colorbox{lightgray}{0.542} & \colorbox{lightgray}{0.657}  & \colorbox{lightgray}{0.698} & \colorbox{lightgray}{0.797} & \colorbox{lightgray}{0.148}\\
% & Ours-full & -/- & - & 0.542 & 0.657  & 0.698 & 0.797 & 0.148\\
\midrule
% \rowcolor{Tan}
% \rowcolor{lightgray}
\multirow{7}{*}{\rotatebox{90}{Token Selection}} & Ours-full & 0.9/0.9 & 1.0 & 0.523 & 0.634 & 0.682 & 0.786 & 0.108\\
% \rowcolor{lightgray}
 & {Ours-full} & \colorbox{Tan}{0.7/0.7} & \colorbox{Tan}{1.0} & \colorbox{Tan}{0.509} & \colorbox{Tan}{0.619} & \colorbox{Tan}{0.671} & \colorbox{Tan}{0.778} & \colorbox{Tan}{0.056}\\
% \rowcolor{Tan}
% \rowcolor{lightgray}
& Ours-full & 0.5/0.5 & 1.0 & 0.432 & 0.571 & 0.596 & 0.743 & 0.028\\
% \hline
\cmidrule{2-9}
% \rowcolor{lightgray}
& Ours-full & 0.7/0.9 & 1.0 & 0.519 & 0.628 & 0.678 & 0.782 & 0.082\\
& Ours-full & 0.9/0.7 & 1.0 & 0.512 & 0.622 & 0.673 & 0.779 & 0.082\\
\cmidrule{2-9}
% \rowcolor{LimeGreen}
% \rowcolor{lightgray}
& Ours-full & 0.7/0.7 & 0.9 & 0.510 & 0.618 & 0.668 & 0.774 & 0.055\\
% \rowcolor{LimeGreen}
% \rowcolor{lightgray}
& Ours-full & 0.7/0.7 & 0.7 & 0.497 & 0.604 & 0.653 & 0.762 & 0.052\\
\bottomrule
\end{tabular}
}
\vspace{-0.4cm}
\end{table}

\begin{figure}[t]
  \centering
%   \fbox{\rule{0pt}{2in} \rule{\linewidth}{0pt}}
   \includegraphics[width=\linewidth]{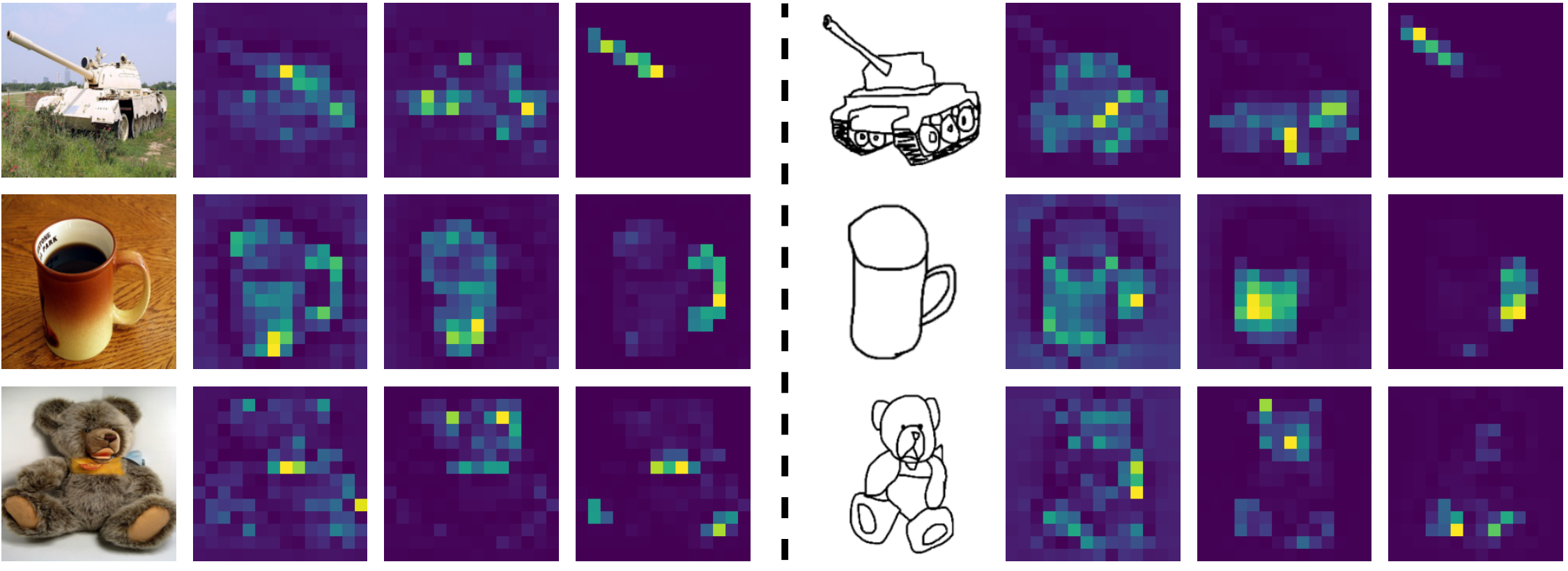}
   \vspace{-0.6cm}
   \caption{Attention maps of self-attention module on unseen categories. Given the tensors (heads) of the last layer of the self-attention module, we display the attention maps by using the retrieval token $[\texttt{Ret}]$ as query. Original inputs are in the first column, followed by attention maps from multiple heads. }
   \label{fig:sa}
   \vspace{-0.7cm}
\end{figure}

\keypoint{Self-attention map.} To investigate what our network has learned from self-attention, we can get self-attention maps by using the retrieval token as query to measure its correlation to each visual token through vector dot-product, similar to \cite{caron2021emerging}.  
As shown in Figure~\ref{fig:sa}, we can observe that different heads can attend to different locations of an image, such as object foreground and different semantic parts, which are useful for fine-grained matching. 

\begin{figure}[t]
  \centering
%   \fbox{\rule{0pt}{2in} \rule{\linewidth}{0pt}}
   \includegraphics[width=\linewidth]{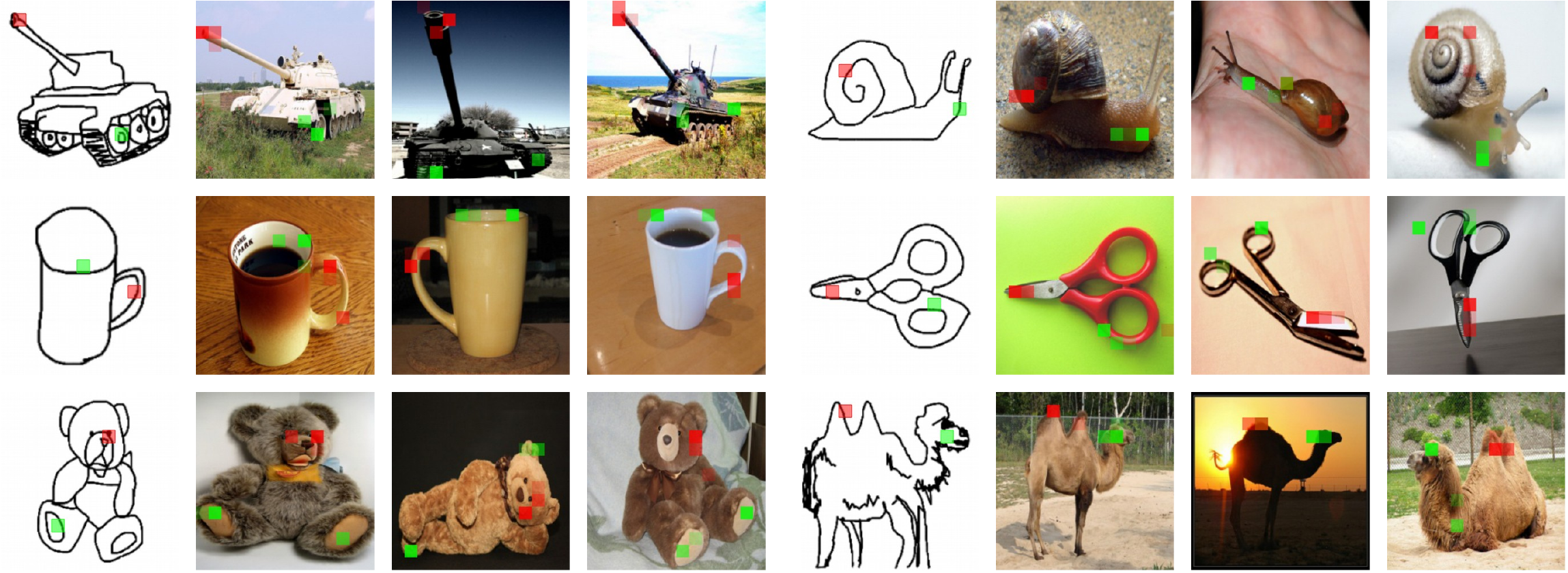}
   \vspace{-0.7cm}
   \caption{Visual correspondence across two modalities. Given a query sketch with two manually selected key regions (color-coded in \textcolor{red}{red} and \textcolor{green}{green}), we show the retrieved images with the corresponding matched regions (Top 3) in the same color.}
   \label{fig:ca}
   \vspace{-0.4cm}
\end{figure}

\keypoint{Cross-modal visual correspondence.} To show our model's capacity of reasoning on salient local regions, we demonstrate some examples of cross-modal local visual correspondence in Figure~\ref{fig:ca}. Specifically, to find estimated correspondences, we can simply measure the vector distance of each visual token pair according to Eq.~\ref{eq:token}. We can clearly see that visual correspondences in the retrieved images can be roughly located given a probe query sketch, despite the objects with different poses and backgrounds, thanks to good robustness of the cross-attention module. 

\begin{figure}[t]
  \centering
%   \fbox{\rule{0pt}{2in} \rule{\linewidth}{0pt}}
   \includegraphics[width=\linewidth]{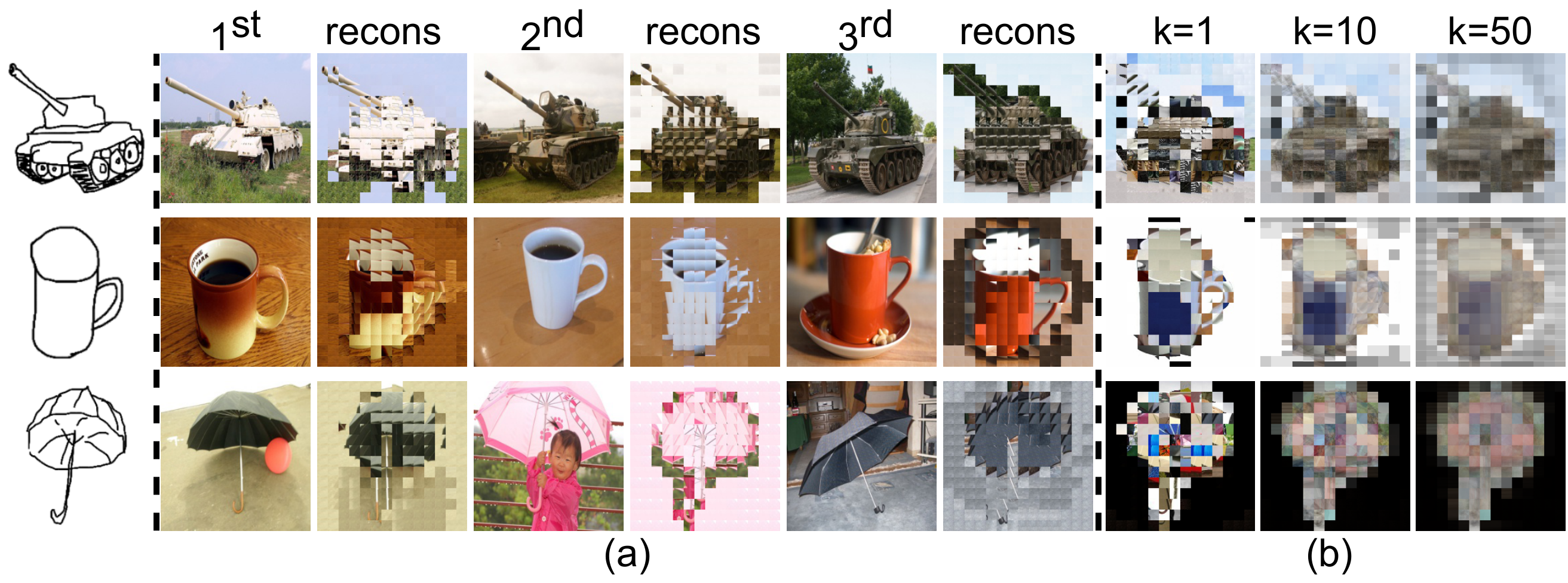}
   \vspace{-0.7cm}
   \caption{Cross-modal patch replacement. Given a sketch, (a) ``recons'' images are obtained by replacing sketch patches with the closest image patches of the top-3 retrieved images. (b) Reconstructed images using the k-nearest patches of the whole gallery. \cut{Zoom in for best view.}}
   \label{fig:recons}
   \vspace{-0.6cm}
\end{figure}

\keypoint{Sketch-to-photo synthesis by patch replacement.} To further inspect if reliable cross-modal correspondence has been discovered, we conduct `sketch-to-photo synthesis' by replacing sketch patches with the closest image patches. The image patches are either found from (i) the retrieved closest image or (ii) the gallery contains all testing images of all categories in Sketchy Ext, i.e., 17,101 images with 3,351,796 patches in our case. Figure~\ref{fig:recons} (a) shows that the synthesized images can resemble the query sketch by stacking semantic closest patches from the retrieved natural images. Interestingly, irrelevant objects in the retrieved images are unselected when doing this sketch-to-photo synthesis, e.g., the \texttt{tray} under the red \texttt{coffee cup}, the \texttt{balloon} and the \texttt{child} under the \texttt{umbrella}. Moreover, besides using the patches from the retrieved images, sketch patches can be replaced by the patch averaged from the top k-nearest image patches from the gallery to investigate the learned correspondence more broadly. As shown in Figure~\ref{fig:recons} (b), the sketch patches can be replaced by quite reasonably similar image patches. Due to a much larger search space, the reconstructed images look scattered when $k=1$. However, meaningful patch-level correspondence still holds, such as the searched \texttt{umbrella handle} and \texttt{tank tracks} patches. While the synthesized images get smoother and more similar content to the query sketches when $k$ is increased. All these results demonstrate the capacity of our model for exploring the local visual correspondences between sketches and images.

\begin{figure}
    \centering
    \includegraphics[width=\linewidth]{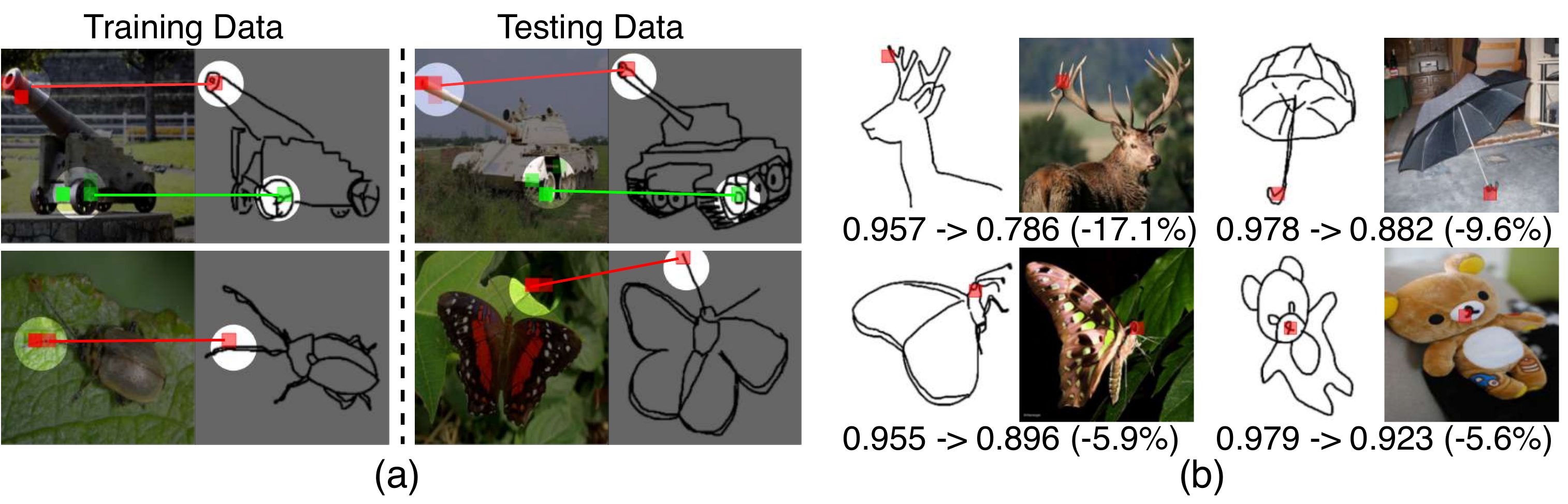}
    \vspace{-0.7cm}
    \caption{(a) Example of shareable local matches. The observed visual correspondences in training data show up again in testing data. (b) Example of most important token pair (\textcolor{red}{red}) which led maximum reduction of the matching score. Zoom in for best view.}
    \label{fig:transfer}
    \vspace{-0.3cm}
\end{figure}

\keypoint{How transfer happens?} To gain some insights about how our model tackles unseen cases after learning local visual correspondence in training data, we take a sketch in test set as query to find the most similar image in training set, then further examine if there are common patterns of local visual matches in both training and testing sketch-photo pairs. As shown in Figure~\ref{fig:transfer} (a), some shareable local matches do exist, such as the \texttt{barrel} and \texttt{wheel} of \texttt{cannon} and \texttt{tank}, suggesting the learned priors of local visual correspondence could be transferred to match regions of novel objects.   

\keypoint{Most influential token pair?} A key issue in explainability of AI systems is to be able to pinpoint the key features of the input that led to a particular decision \cite{samek2019explainable}. Various off-the-shelf methods \cite{selvaraju2017gradCam} exist for this in recognition systems, but it is trickier for retrieval systems as decisions operate on pairs of inputs. Our method provides the ability to answer such questions by identifying the most important feature pairing responsible for a match. Specifically, we remove one pair at a time, and return the pair that leads to maximum reduction of relation score. Some examples are demonstrated in Figure~\ref{fig:transfer} (b). We can see that, for example, the most influential token is the \texttt{antler} of \texttt{deer}, which led a maximum reduction about $17\%$ of matching score.

\keypoint{Computational cost analysis.}  
We compare the GFLOPs, model size and runtime per pair of sketch-image matching (RPM) between ours and two SOTA methods, i.e., SAKE~\cite{liu2019semantic} and SEM-PCYC~\cite{wang2021domain}. From Table~\ref{tab:flops}, we can see our model has fewer parameters than SEM-PCYC. In terms of GFLOPs and RPM, it is dominated by the self-attention module, i.e. the core module of the used ViT backbone. However, the employed token selection can reduce the GFLOPs and RPM significantly while still deliver the SOTA performance, i.e., Ours$\star$ in Table~\ref{tab:flops} which is the same model variant color-coded in brown in Table~\ref{table:abla}. More importantly, it is worth noting that our proposed CA module causes only a modest cost ($8.7\%$ of GFLOPs and $14.1\%$ of parameters) to achieve a significant performance gain, C.F. w/o CA vs Ours-full shown in Table~\ref{table:abla}.

\setlength{\tabcolsep}{1pt}
\begin{table}[t]
\small
    \caption{Comparison of computational cost.}
    \vspace{-0.3cm}
    \label{tab:flops}
    \centering
    \scalebox{0.75}{
    \begin{tabular}{l|ccccc}
    \toprule
        
         & SAKE \cite{liu2019semantic} & SEM-PCYC \cite{dutta2019semantically} &  Ours-RN (SA+CA) & Ours$\star$ \\
        \midrule
        \# Params (M) & 27.6 & 137.9 & 102.2(87.8+14.4) & 102.2 \\
        GFLOPs & 3.90 & 15.5 & 19.5 (17.8+1.7) & 12.6 (12.0+0.6) \\
        RPM (ms) & 0.138 & 0.070 & 0.148 (0.118+0.030) & 0.056 (0.048+0.008) \\
        \bottomrule
    \end{tabular}}
    \vspace{-0.3cm}
\end{table}

\subsection{Zero-shot Fine-grained SBIR}

\keypoint{Datasets.} 
QMUL-Shoe-V2 and QMUL-Chair-V2 \cite{yu2016sketch} are two commonly used benchmarks for FG-SBIR. QMUL-Shoe-V2 is composed of 6,730 shoe sketches with 2,000 associated photos, and QMUL-Chair-V2 contains 2,000 chair sketches with the corresponding 400 photos. We train our model on QMUL-Shoe-V2 then test on QMUL-Chair-V2 to conduct ZS FG-SBIR experiments.

\keypoint{Evaluation protocol.} We follow the evaluation protocol  in \cite{sain2021stylemeup}, i.e., metrics of top 1 and top 10 retrieval accuracy. I.e., credit will be given if the true positive (photo) to the query (sketch) is ranked within the top 1/10 slots.

\keypoint{Competitors.} Five strong FG-SBIR baselines are compared, including TripLet-SAN \cite{yu2016sketch}, DSA \cite{song2017deep}, TripLet-RL \cite{bhunia2020sketch}, CC-DG \cite{pang2019generalising}, and StyleMeUp \cite{sain2021stylemeup}.  \cut{TripLet-SAN \cite{yu2016sketch} adopts Sketch-A-Net \cite{yu2017sketch} as network backbone and is trained by using triplet loss. DSA \cite{song2017deep} leverages spatial attention with a higher order HOLEF ranking loss for FG-SBIR. TripLet-RL \cite{bhunia2020sketch} adopts a reinforcement learning approach for training an early-retrieval agent. For fair comparison, only complete sketches are used as queries here. CC-DG \cite{pang2019generalising} is a cross-category domain generalization method for FG-SBIR. CC-DG aims to learn a shared manifold of prototypical visual sketch traits, which are used to re-parameterize the feature extractor when solving unseen categories. StyleMeUp \cite{sain2021stylemeup} proposes a style-agnostic SBIR method which disentangles sketch/photo features into a cross-modal-invariant semantic space and a drawing style relevant space. Retrieval then can be conducted such that the variations of drawing style are ignored. }

\keypoint{Results.} From Table~\ref{tab:fg-sbir}, we can see that our approach surprisingly surpasses all baselines even under an \textit{unfair} comparison, i.e., ours is tested in a zero-shot setting, whereas all competitors are trained on the target category.

\setlength{\tabcolsep}{3pt}
\begin{table}[]
\small
    \centering
    \caption{Zero-shot FG-SBIR results ($\%$). {Note that all competitors are \emph{not} zero-shot models, they are trained on Chair-V2.}}
    \vspace{-0.2cm}
    \label{tab:fg-sbir}
    \scalebox{0.8}{
    \begin{tabular}{lccc}
        \toprule
        \bf Method & TripLet-SAN \cite{yu2016sketch} & DSA \cite{song2017deep} & TripLet-RL \cite{bhunia2020sketch} \\
        \midrule
          acc.@1 & 47.65 & 53.41 & 56.54 \\
          acc.@10 & 84.24 & 87.56 & 89.61 \\ 
          \midrule
          \bf Method & StyleMeUp \cite{sain2021stylemeup} & CC-DG \cite{pang2019generalising} & Ours-RN/Ours-Ret \\
          \midrule
          acc.@1 & 62.86 & 54.21 & 63.34/\textbf{64.31}\\
          acc.@10 & 91.14 & 88.23 & \textbf{94.53}/92.60\\
         \bottomrule
    \end{tabular}
    }
    \vspace{-0.5cm}
\end{table}

\setlength{\tabcolsep}{1pt}
\begin{table}
\small
% \vspace{-0.2cm}
\begin{center}
\caption{Cross-dataset ZS-SBIR results. ``S'', ``T'' and ``Q'' denote Sketchy Ext, TU-Berlin Ext, and QuickDraw Ext, respectively.  ``$(\cdot)$'' denotes the number of test categories which are unseen to ensure the zero-shot setting. E.g., S$\to$T(21) denotes that, we train on the training split of Sketchy Ext, then test on a subset (21 unseen classes) of the testing split of TU-Berlin Ext. Rows with a grey background indicate using ViT backbone for fair comparisons.}
% \vspace{-0.3cm}
\label{table:open-sbir}
\scalebox{0.75}{
\begin{tabular}{lcccccccc}
\toprule
\multirow{2}{*}{\bf Method} & \multicolumn{2}{c}{\bf S$\to$ T (21)} & \multicolumn{2}{c}{\bf S$\to$ Q (11)} & \multicolumn{2}{c}{\bf T$\to$ S (8)} & \multicolumn{2}{c}{\bf T$\to$ Q (10)}\\ 
\cline{2-9}
& mAP & Prec@100  & mAP & Prec@100 & mAP & Prec@100  & mAP & Prec@100 \\
\midrule
\multirow{2}{*}{CC-DG \cite{pang2019generalising}}  & 0.252 & 0.403  & 0.148 & 0.212 & 0.570 & 0.660  & 0.214 & 0.278 \\
 &  \colorbox{lightgray}{0.308} & \colorbox{lightgray}{0.434} & \colorbox{lightgray}{0.156} & \colorbox{lightgray}{0.227} & \colorbox{lightgray}{0.624} & \colorbox{lightgray}{0.693}  & \colorbox{lightgray}{0.231} & \colorbox{lightgray}{0.296} \\
\multirow{2}{*}{DSN \cite{wang2021domain}}  & 0.384  & 0.480 & 0.152 & 0.171 & 0.646 & 0.673 & 0.229 & 0.251 \\
 &  \colorbox{lightgray}{0.356} & \colorbox{lightgray}{0.469} & \colorbox{lightgray}{0.149} & \colorbox{lightgray}{0.178} & \colorbox{lightgray}{0.613} & \colorbox{lightgray}{0.654} & \colorbox{lightgray}{0.218} & \colorbox{lightgray}{0.246} \\
\multirow{2}{*}{SAKE \cite{liu2019semantic}}  & 0.421 & 0.549  & 0.183 & 0.250 & 0.657 & 0.722 & 0.248 & 0.340 \\
 &  \colorbox{lightgray}{0.389} & \colorbox{lightgray}{0.506} & \colorbox{lightgray}{0.174} & \colorbox{lightgray}{0.242} & \colorbox{lightgray}{0.626} & \colorbox{lightgray}{0.701} & \colorbox{lightgray}{0.235} & \colorbox{lightgray}{0.318} \\
\hline
Ours-RN & \textbf{0.476} & \textbf{0.590}  & \textbf{0.228} & \textbf{0.338} & \textbf{0.746} & \textbf{0.816} & \textbf{0.273} & \textbf{0.376} \\
\bottomrule
\end{tabular}
}
\end{center}
\vspace{-0.7cm}
\end{table}

\subsection{Cross-Dataset category-level ZS-SBIR}
% \vspace{-0.1cm}
\keypoint{Settings.} We finally verify the ability of our method to generalize across completely different datasets in zero-shot scenario, i.e., the model is trained on dataset A then tested on dataset B, where the test classes are all unseen during training. This setting goes beyond the standard within-dataset ZS-SBIR benchmarks to evaluate transfer across entirely different benchmarks. Such cross-dataset ZS-SBIR is an even more challenging and realistic setting, since sketches from different datasets are typically drawn in diverse styles, as well as containing disjoint classes. 

\keypoint{Results.} As shown in Table~\ref{table:open-sbir}, our method works much better than other competitors, demonstrating the preferable generalization ability of our learned visual correspondences over both the dataset and task shifts.

% \vspace{-0.2cm}
\section{Conclusion and Future Work}
% \vspace{-0.2cm}
We tackled ZS-SBIR, with a hope to also make it explainable. We are inspired by ``old vision'', and put forward a patch matching framework, that is not only explainable but also able to tackle all ZS-SBIR settings at the same time. The technical solution is a transformer-based cross-modal network, with three specific designs to tailor to the problem: (i) a self-attention module to learn the tokens, (ii) a cross-attention module to establish putative matches, and (iii) {a kernel-based relation network to aggregate local matches into an overall similarity score. Last but not least, visualizations on patch-level correspondences, and sketch-to-photo synthesis through cross-modal patch replacement, provide means of explanation.} 

We can see 
the false matchings in Figure~\ref{fig:retrieval} are caused by the high similarity either between local tokens, e.g., windmill's paddles v.s. scissors' blades or global shapes, e.g., pizza v.s. fan. Therefore, how to take the best of global and local correspondence is worth exploring. Moreover, our patch replacement-based photo synthesis is rather coarse, thus improving synthesis quality will be a future endeavour.

\clearpage
\newpage
%%%%%%%%% REFERENCES
{\small
\bibliographystyle{ieee_fullname}
\bibliography{main}
}

\end{document}